С. Н. Бук

# РОМАН ІВАНА ФРАНКА “ДЛЯ ДОМАШНЬОГО ОГНИЩА” КРІЗЬ ПРИЗМУ ЧАСТОТНОГО СЛОВНИКА

У статті описано методику та принципи укладання частотного словника роману І. Франка “Для домашнього огнища”. Отримано статистичні параметри лексики твору, зокрема, індекси різноманітності (багатство словника), винятковості, концентрації, співвідношення між рангом слова та величиною покриття тексту тощо. На основі частотних словників романів І. Франка “Перехресні стежки” (2007) та “Для домашнього огнища” (2010) здійснено порівняння основних кількісних характеристик вказаних текстів.

## 1. Вступ

Частотність лексичної одиниці — це важлива характеристика слова, оскільки вона свідчить про активність його функціонування в тексті, про його вагу у статистичній структурі тексту тощо. На цьому наголошують як українські, так і зарубіжні мовознавці: В. Перебийніс[1], Н. Дарчук[2], Т. Грязнухіна[3] М. Муравицька[4], Ґ. Віммер[5], Ґ. Альтманн[6], І. Попеску[7], Р. Кьолер[8], Ю. Тулдава[9], П. Алесеєв[10], А. Шайкевич[11], А. Павловскі[12], М. Рушковскі[13] та ін.

З'явилась і нова галузь лінгвістичної науки – статистична лексикографія, що, зокрема, займається теорією та практикою укладання частотних словників. Такий інтерес до статистичного обстеження великих сукупностей текстів зумовлений, з одного боку,

---

1 *Перебийніс В. С.* Статистичні методи для лінгвістів: Навчальний посібник.— Вінниця: “Нова книга”, 2002.— 168 с.; *Перебийніс В. С.* Частотний словник // Українська мова: Енциклопедія / редкол.: *В. М. Русанівський* та інші.— К.: Українська енциклопедія, 2000.— С. 790.; *Перебийніс В. С., Муравицька М. П., Дарчук Н. П.* Частотні словники та їх використання.— К: Наукова думка, 1985.— 204 с.; Статистичні параметри стилів / за ред. *В. С. Перебийніс*.— К.: Наук. думка, 1967.— 260 с.

2 *Дарчук Н. П.* Статистичні характеристики лексики як відображення структури тексту // Мовознавчі студії.— К.: Наукова думка, 1976.— С. 97–102.

3 *Дарчук Н. П., Грязнухіна Т. О.* Частотний словник сучасної української публіцистики // Мовознавство.— 1996.— №4–5.— С. 15–19.

4 *Муравицька М. П.* Статистичні лінгвістичні дослідження та їх розвиток в українському мовознавстві // Мовознавство.— 1967.— № 5.— С. 47–58.

5 *Wimmer W., Altmann G.* Review Article: On vocabulary richness // Journal of Quantitative Linguistics 1999. Vo. 6. No. 1. P 1-9.

6 *Wimmer G., Altmann G., Hřebíček. L., Ondrejovič, S., Wimmerová. S.* Úvod do analýzy textov.— Bratislava: Veda, 2003.— 344 s.

7 *Popesku I.-I., Altmann G.* Some aspects of word frequencies // Glottometrics.— 13.— 2006.— P. 23-46

8 *Köhler R., Altmann G.* Aims and Methods of Quantitative Linguistics // Problems of Quantitative Linguistics. — Chernivci: Ruta, 2005.— P. 12–42.

9 *Тулдава Ю. П.* Проблемы и методы квантитативно-системного исследования лексики.— Таллин: Валгус, 1987.— С. 62–64.

10 *Alekseev P. M.* Frequency dictionaries // Quantitative Linguistik : ein internationales Handbuch = Quantitative linguistics : an international handbook / edited by *Reinhard Köhler, Gabriel Altmann, Rajmund G. Piotrowski*. — Berlin ; New York: Walter de Gruyter, 2005. — P. 312-324.; *Алексеев П. М.* Частотные словари: Учебное пособие.— СПб.: Изд-во С.-Петерб. ун-та, 2001.— 156 с.

11 *Шайкевич А. Я.* О статистическом словаре языка Достоевского // Русский язык в научном освещении.— М., 2001.— № 2.— С. 122–149.; *Шайкевич А. Я.* Количественные методы в языкознании // Лингвистический энциклопедический словарь / ред. *Н. Л. Ярцева*.— М.: Советская энциклопедия, 1990.— С. 231–232.

12 *Pawłowski A.* Metody kwantytatywne w sekwencyjnej analizie tekstu.— Warszawa: UW, 2001.— 168 s.

13 *Ruszkowski M.* Statystyka w badaniach stylistyczno-składniowych.— Kielce: Wydawnictwo Świętorszyskiej, 2004.— 144 s.

необхідністю глибше пізнати закономірності функціонування мовних одиниць у тексті. З іншого боку, виникнення статистичної лексикографії зумовлене суспільними потребами, практичним застосуванням надбань мовознавства для удосконалення системи стенографії і відбору лексичного мінімуму для словників різного типу, раціонального використання пам'яті комп'ютера для автоматичного опрацювання текстової інформації, теорії та практики перекладознавства[14].

Загальне застосування і ціль частотного словника (ЧС) полягає в опрацюванні ймовірнісно-статистичної моделі системного лінгвістичного об'єкту, а саме мови (підмови, стилю, ідіолекту)[15]. Проте насамперед основне завдання ЧС — це теоретичні дослідження мови, на базі якої створено частотний словник. Серед тенденцій та перспектив розвитку статистичної лексикографії П. Алексеєв вказує на залучення до дослідження величезної кількості мов та підмов, перехід від загального до часткового. Кількісно-статистичний опис ідіолекту письменника природно входить у цей контекст.

Таким чином, на стику авторської та статистичної лексикографії виник частотний словник мови письменника, і це закономірно, оскільки для опису ідіолекту письменника необхідні не лише якісні, а й кількісні характеристики лексики його творів. Саме тому для найвидатніших майстрів пера різних часів та народів укладали частотні словники, де наводили кількість вживань слова в обстежених текстах певного обсягу. Так, існують ЧС Арістотеля[16], В. Шекспіра[17], В. Гюго[18], Дж. Джойса[19], К. Чапека[20], Ф. Достоєвського[21], Л. Толстого[22], А. Чехова[23], М. Лермонтова[24], М. Горького[25], болгарського поета Н. Вапцарова[26], польського письменника К. Бачинського[27], сербських прозаїків М. Павича[28], Д. Албахарія[29], М. Данойліча[30], С. Велмар-Янковіч[31] тощо.

В українській лексикографії інформацію про частоту вживання слова містять

---

14 *Rybicki Jan* Stylometria komputerowa w służbie tłumacza (na przykładzie własnych przekładów) // Rocznik przekładoznawczy: Studia nad teorią, praktyką i dydaktyką przekładu.— 2007-2008.— № 3/4.

15 *Alekseev P. M.* Frequency dictionaries ...

16 *Colin, B. Rutten, C.* (1993), Aristote, categoriae: index verborum, listes de fréquence. Liège: Université de Liège.

17 Complete Shakespeare wordlist. Available from <http://en.wiktionary.org/wiki/Wiktionary:Frequency_lists/Complete_Shakespeare_wordlist>

18 *Brunet, E.* (1988), Le vocabulaire de Victor Hugo. Genève: Slatkine.

19 *Hanley, M. L.* (1951), Word index to James Joyce's "Ulysses". Madison: University of Wisconsin Press.; *Ulysses by J. Joyce* A Ranked Concordance <http://www.doc.ic. ac.uk/~rac101/concord/texts/ulysses/>

20 Slovník Karla Čapka / Hl. ed. František Čermák. — Praha : Nakladatelství Lidové noviny ; Ústav Českého národního korpusu, 2007. — 715 s.

21 *Шайкевич А. Я., Андрющенко В. М., Ребецкая Н. А.* Статистический словарь языка Достоевского.— М.: Языки славянской культури, 2003.— 880 с.

22 *Великодворская З. Н. и др.* (ред.) Частотный словарь романа Л. Н. Толстого "Война и мир" / М-во просвещения РСФСР.— Тула: Б. и., 1978.

23 *Гребенщиков А. О.* Частотный словарь рассказов А. П. Чехова / под ред. *Г. Я. Мартыненко*.— СПб: Изд-во Снкт-Петербург. ун-та, 1999.

24 Частотный словарь языка М. Ю. Лермонтова // Лермонтовская энциклопедия.— М: Советская энциклопедия, 1981.— С. 717–774.

25 *Алексеев П.* Частотный словарь автобиографической трилогии М. Горького.— Санкт-Петербург: Изд-во Санкт-Петербугского ун-та, 1996.— 205 с.

26 Честотен речник на Вапцаровата поезия / Авт. кол.: *Г. В. Крилова, А. А. Азарова, Е. А. Захаревич* и др.— Велико Търново: Абагар, 1996.

27 *Balowski M.* Lista frekwencyjna poezji, prozy i dramatu Krzystofa Kamila Baczyńskiego.— Prohowice: Pro, 1997.

28 *Vasić S.* Polazne osnove novije srpske proze. Knjiga II. Hazarski rečnik Milorada Pavića: frekvencijski rečnik.— Beograd: Kultura, 1998.

29 *Vasić S.* Polazne osnove novije srpske proze — frekvencijski rečnik romana Mamac Davida Albaharija.— Beograd: Kultura, 2001.

30 *Vasić S.* Polazne osnove novije srpske proze — frekvencijski rečnik romana Oslobodioci i izdajnici Milorada Danojlića.— Beograd: Kultura, 2002.

31 *Vasić S.* Polazne osnove novije srpske proze — frekvencijski rečnik romana Bezdno Svetlane Velmar-Janković.— Beograd: Kultura, 2000.

словники мови Т. Шевченка[32] та Г. Квітки-Основ'яненка[33], словопокажчики до творів І. Котляревського[34], Т. Шевченка[35], Лесі Українки[36], Ю.Федьковича[37], А. Тесленка[38] та інших письменників.

Електронний частотний словник сучасної української поетичної мови доступний в Інтернеті[39]. Його вибіркою (обсягом 300 тис. слововживань) стали твори таких українських митців ХХ ст., як Ю. Андрухович, М. Вінграновський, М. Воробйов, І. Драч, І. Жиленко, О. Забужко, С. Йовенко, Т. Коломієць, Л. Костенко, І. Малкович, Б. Олійник, Д. Павличко, В. Стус, Л. Скирда.

Перші засади словникового опису творів І. Франка заклав І. Ковалик[40], зокрема вийшла друком "Лексика поетичних творів Івана Франка"[41], що подає усі лексеми — 35 000 — поезії письменника за алфавітом із вказівкою на частотність кожної з них. На основі цієї праці виконано низку досліджень, наприклад, статистичний аналіз лексики[42], кількісний аналіз антропонімів[43] тощо.

Зараз реалізовується проект комплексного квантитативного опису творів І. Франка[44], що розпочався з етапу кількісної параметризації великої прози письменника. Він передбачає поетапне укладання ЧС романів (повістей) автора (зокрема, вийшов друком ЧС роману "Перехресні стежки"[45]), з подальшим їх зіставленням і виявленням багатства Франкового мовлення, особливостей добору мовних засобів, підтвердженням чи спростуванням статистичних законів, що також вносить уточнення до ідіолекту письменника.

Заплановано, що ЧС І. Франка дає кількісну характеристику одиницям мовлення письменника, а тим самим детальний та вичерпний перелік лексичних одиниць, що може стати базою для укладання словника його мови тлумачного типу.

---

32 Словник мови Шевченка: У 2 т. / *Ващенко В. С.* (ред.).— К.: Наукова думка, 1964.

33 Словник мови творів Г. Квітки-Основ'яненка: У 3 т. / Відп. ред. *М. А. Жовтобрюх.*— Х.: Харків. держ. ун-т, 1978.

34 *Ващенко В. С., Медведєв Ф. П., Петрова П. О.* Лексика "Енеїди" І. Котляревського. Покажчик слововживання.— Х.: В-во Харківського ун-ту, 1955.— 207 с.; *Бурячок А. А., Залишко А. Т., Ротач А. О., Северин М. Д.* Лексика п'єс та од І. П. Котляревського / За ред. *А. Бурячка.*— К.: Вища школа, 1974.— 54 с.

35 *Ващенко В. С., Петрова П. О.* Шевченкова лексика. Словопокажчик до поезій Т. Г. Шевченка.— К.: Видавництво Київського державного університету ім. Т. Шевченка, 1961.— 106 с.

36 Бойко М. Ф. Словопокажчик драматичних творів Лесі Українки.— К.: Видавництво АН УРСР, 1961.— 93 с.

37 *Лук'янюк К. М.* (гол. ред.) Юрій Федькович: Словопокажчик мови творів письменника.— Чернівці: Місто, 2004.— 188 с.

38 *Сизько Т.* Лексика мови Архипа Тесленка. Словопокажчик оповідань.— Дніпропетровськ, 1970.— 101 с.

39 Частотний словник сучасної української поетичної мови (2003-2008) <www.mova.info/Page2.aspx?l1=89>

40 *Ковалик І. І.* Принципи укладання Словника мови творів Івана Франка // Українське літературознавство. Іван Франко. Статті і матеріали.— Львів, 1968.— Вип. 5.— С. 174–183.; *Ковалик І. І.* Наукові філологічні основи укладання і побудови Словника мови художніх творів Івана Франка // Українське літературознавство. Іван Франко. Статті і матеріали.— Львів, 1972.— Вип. 17.— С. 3–10.; *Ковалик І. І.* Словник мови художніх творів Івана Франка. Пробний зошит // Українське літературознавство. Іван Франко. Статті і матеріали.— Львів, 1976.— Вип. 26.— С. 63–99.

41 *Ковалик І. І., Ощипко І. Й., Л. М. Полюга* (уклад.) Лексика поетичних творів Івана Франка. Методичні вказівки з розвитку лексики.— Львів: ЛНУ, 1990.— 264 с.

42 *Полюга Л.* Статистичний аналіз лексики поетичних творів І. Франка // Іван Франко і національне відродження.— Львів: ЛДУ ім. І Франка; Ін-т франкознавства, 1991.— С. 164–166.

43 *Полюга Л.М.* До питання про опрацювання української літературної ономастики // Ономастика і апелятиви: Зб. наук. пр. – Дніпропетровськ, 1998. – С. 42–47.

44 *Бук С.* Квантитативна параметризація текстів Івана Франка: спроба проекту // Іван Франко: Студії та матеріали: Збірник наукових статей.— Львів: ЛНУ імені Івана Франка, 2010. *(у друці)* (у друці), див. препринт arXiv:1005.5466v1 [cs.CL] за адресою <http://arxiv.org/abs/1005.5466>.

45 *Бук С., Ровенчак А.* Частотний словник повісті І. Франка "Перехресні стежки" // Стежками Франкового тексту (комунікативні, стилістичні та лексикографічні виміри роману "Перехресні стежки") / *Ф. С. Бацевич* (наук. ред). – Львів: Видавничий центр ЛНУ імені Івана Франка, 2007.— С. 138-369.

## 2. Місце роману "Для домашнього огнища" у творчості І. Франка

Роман "Для домашнього огнища" займає важливе місце у творчості І. Франка. Це єдиний великий прозовий твір І. Франка, написаний не від номера до номера, а відразу увесь від першого до останнього рядка[46]. Автор створив його спершу польською мовою, перебуваючи у Відні на славістичних студіях В. Ягича і готуючись до захисту докторату. В. Щурат, що мешкав тоді разом з ним, згадує: "... Франко творив при мені повість "Для домашнього огнища" на основі голосного у Львові судового процесу. Міркуючи, що повість може бути для поляків цікавіша ніж для українців, написав її зразу по-польський зимою 1892 та й передав її до Варшави через Зигмунда Василевського ..."[47].

У листі до Я. Карловича І. Франко пише: "Я, звичайно, як добрий батько, хотів би це моє дітище добре продати тим більше, що, залишивши роботу у "Kurjer'i Lw[owskim]" і готуючись до захисту докторату, потребую грошенят як для свого перебування у Відні, так і для утримання дружини і дітей у Львові"[48], далі просить подати твір до декількох варшавських редакцій, щоби отримати "дещо більший аванс (30-50 %)"[49]. Вважається, що саме гостра потреба у грошах змусила І. Франка написати цей твір по-польськи — зі сподіванням надрукувати його у варшавському виданні, де можна розраховувати на "дещо більший аванс"[50]. Проте польською мовою твору тоді так надруковано й не було: не пройшов цензури.

У вересні 1893 І. Франко почав перекладати текст українською. Оскільки "зміна мови передбачає зміну читацької аудиторії, ... певну зміну усього читацького поля загалом"[51], то й перекладав роман автор не дослівно, а переорієнтовуючись на українського читача з його характерними смаками. Насамперед написав присвяту О. Мирному, створюючи алюзії на його "Повію". Україномовний твір вперше повністю надруковано 1897 у серії "Літературно-наукова бібліотека", кн. 11.

Роман піднімає важливу проблему суспільства[52], яка не втратила актуальності й сьогодні. Головна героїня Анеля Ангарович, опинившись із двома дітьми без засобів для існування, коли чоловік-військовий був на війні в Боснії, стає частиною "бізнесу", що займається вербуванням-продажем молодих дівчат у публічні будинки. 1992 р. за сюжетом твору знято художній фільм із однойменною назвою (режисер Борис Савченко).

## 3. Методика укладання частотного словника роману І. Франка "Для домашнього огнища"

ЧС роману укладено за виданням *Франко І.* Для домашнього огнища // Зібрання творів у 50-ти томах.— Т. 19: Повісті та оповідання.— К.: Наукова думка, 1979.— С. 7–143 із врахуванням прижиттєвого видання *Франко І.* Для домашнього огнища: Повість Івана Франка.— Львів: З друкарні Інститута Ставропігійського. Під зарядом Йосифа Гузара, 1897.— 149 с. (Літературно-наукова Бібліотека. Нова серія; Кн 11).

### 3.1. Зіставлення видань твору 1979 і 1897 рр.

Відомо, що в другій половині XIX ст. існували різні правописні системи, але жодна з них не була обов'язковою[53]. Тому в процесі роботи над ЧС роману "Для домашнього

---

46 *Мороз М.* "Для домашнього огнища" (творча та видавнича історія повісті І. Франка) // Радянське літературознавство.— 1982.— № 4.— С. 30.

47 *Щурат В.* Франків спосіб творення // Спогади про Івана Франка / Упор., вступ. ст. і примітки М. Гнатюка.— Львів: Каменяр, 1997.— С. 281.

48 Тут і далі переклад наш.

49 *Франко І.* До Яна Карловича // Зібрання творів у 50-ти томах.— Т. 49.— К.: Наукова думка, 1979.— С. 368

50 *Мороз М.* Цит.праця. С. 29-30; *Тодчук Н.* Роман І. Франка "Для домашнього огнища": простір і час.— Львів, 2002.— С. 14.

51 *Тодчук Н.* Цит.праця. С. 27.

52 *Вервес Г. Д.* Проблематика польських повістей Івана Франка, "Радянське літературознавство", 1963, № 3, с. 64–82.

53 *Терлак З.* Мовно-правописне редагування творів Івана Франка і проблема збереження автентичності авторського тексту // Іван Франко: дух, наука, думка, воля: Матеріали Міжнародного наукового конґресу, огнища” виявлено основні відмінності між текстами вказаних видань. Так, відновлено написання літери “ґ” у словах ґудзик, ґешефт, цуґ, Боґа мі! (сербськ. Боже мій), ґетівський, леґейда. На відміну від прижиттєвого видання “Перехресних стежок”[54], через літеру “г”, а не “ґ” написано слова грунт, гвалт, морг (міра площі) та ін. Можна було очікувати фарингальне “ґ” у власних назвах Рейхлінген, Големб'я, проте у першодруці використано глоткове “г”.

Також виявлено значну кількість інших правописних відмінностей, які, проте, не впливають на статистичну структуру тексту на лексичному рівні: використання на місці сучасної “ї” після голосних літери “і” або буквосполуки “йі” (*єі, боіть ся, йій, прийіде, вспокоіти(ся), твоім, другоі, приізду, Боснії*); написання “і” на місці деяких сучасних “и” (*сістематично, трівожились, трівоги, корідора*) та в Р. в. однини іменників ж.р. (*памяти, осени, хвили*); використання суфікса та закінчення Н. в. іменників середнього роду -н(є), -т(є), -с(є) на місці сучасного -нн(я), -тт(я), -сс(я) (*чутє, прочутє, бажанє, розчарyванє, привитанє, волосє але зворушеня*) та інші.

Зафіксовано також відмінності, які впливають на кількісні характеристики лексем: написання зворотної дієслівної частки -ся окремо (а -сь — разом), слова “пів” та частки “-небудь” — окремо (*пів години, що небудь*), частки “ж” — разом (*робиж*), складених прислівників типу *раз-у-раз* — через дефіс; написання сучасних прислівників, що починаються колишніми прийменниками, окремо (*по перед, з близька, на ліво, що хвиля , з далека, до дому*), тощо. Цікаво, що частка *-ся* у творі трапляється 1285 разів, це друге (!) за частотою значення після І / Й.

Привертає увагу, що в оригінальному тексті польське прізвище Szablińska фонетично наближене до української вимови: Шаблинська, тоді як у виданні 1979 р. подано Шаблінська, а ім'я головної героїні Aniela в оригіналі наближене до польського звучання: Анєля, напротивагу в 50-томнику — Анеля.

За винятком вказаних різнописань тексти видань загалом ідентичні.

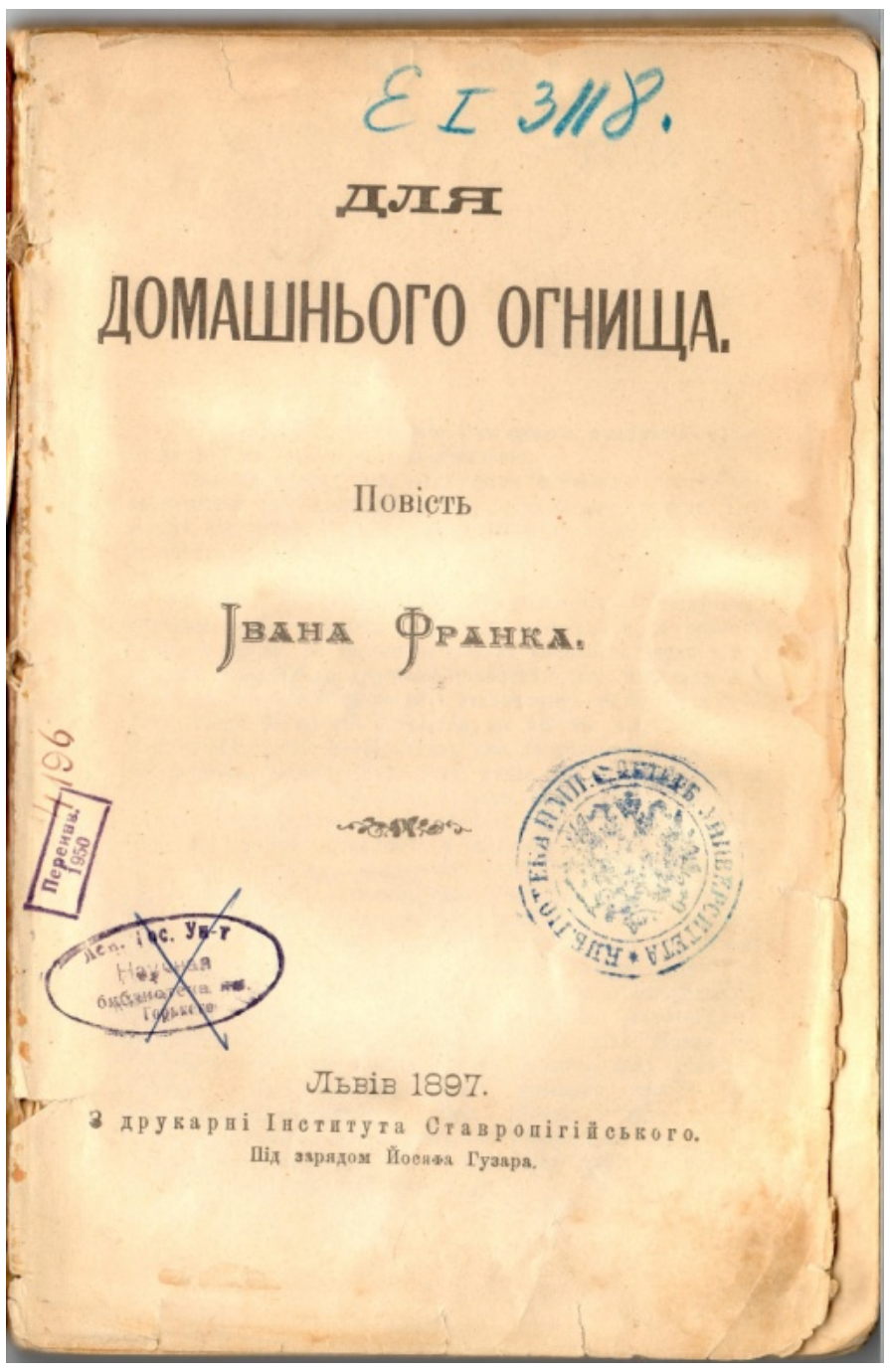

Є I 3118.

ДЛЯ

ДОМАШНЬОГО ОГНИЩА.

Повість

Івана Франка.

Львів 1897.

З друкарні Інститута Ставропігійського.

Під зарядом Йосифа Гузара.

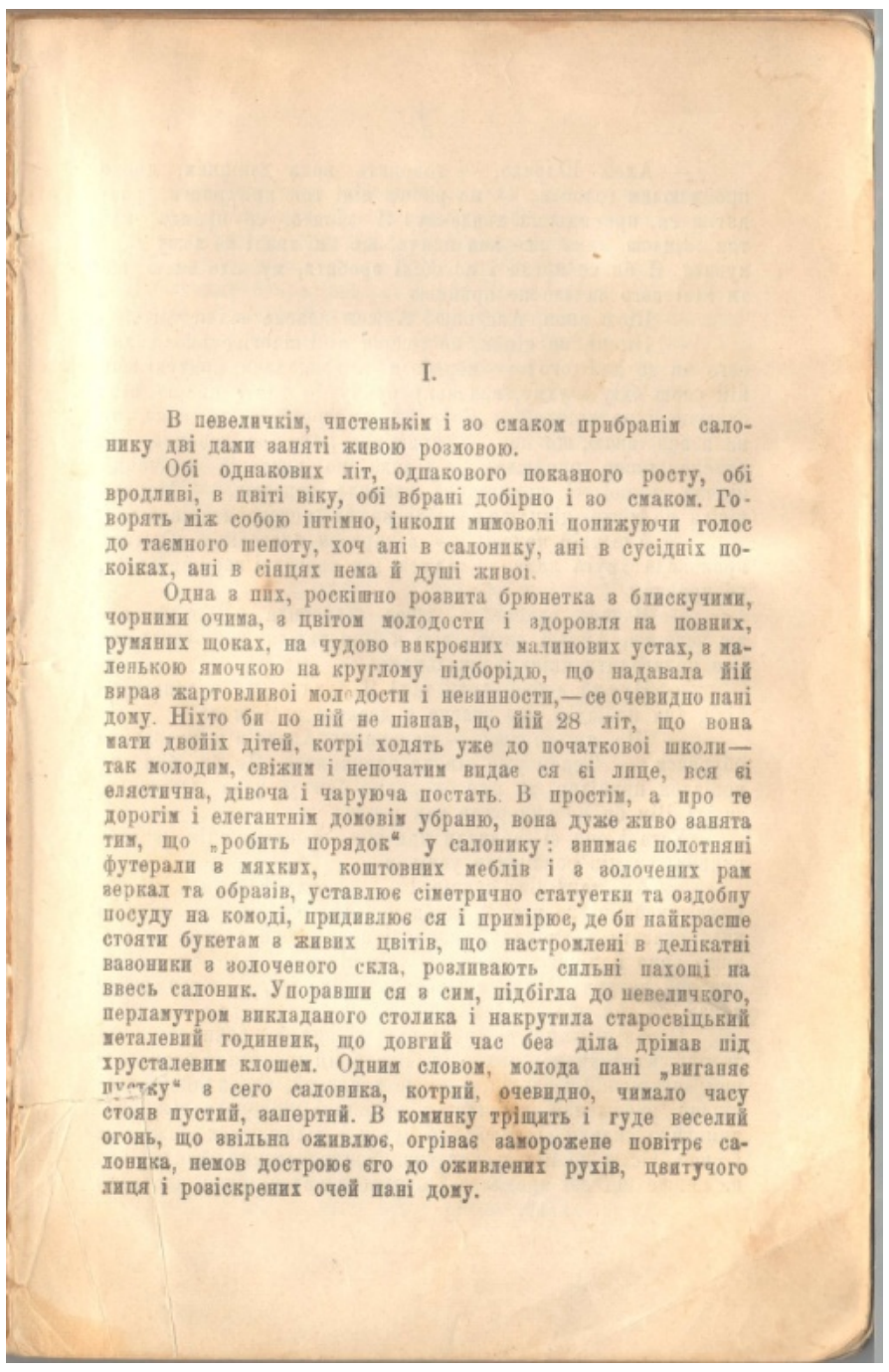

I.

В невеличкім, чистенькім і зо смаком прибранім салонику дві дами заняті живою розмовою.

Обі однакових літ, однакового показного росту, обі вродливі, в цвіті віку, обі вбрані добірно і зо смаком. Говорять між собою інтімно, інколи мимоволі понижуючи голос до таємного шепоту, хоч ані в салонику, ані в сусідніх покоіках, ані в сінцях нема й душі живоі.

Одна з них, роскішно розвита брюнетка з блискучими, чорними очима, з цвітом молодости і здоровля на повних, румяних щоках, на чудово вкроєних малинових устах, з маленькою ямочкою на круглому підборідю, що надавала йій вираз жартовливоі молодости і невинности, — се очевидно пані дому. Ніхто би по ній не пізнав, що йій 28 літ, що вона мати двойіх дітей, котрі ходять уже до початковоі школи — так молодим, свіжим і непочатим видає ся єі лице, вся єі елястична, дівоча і чаруюча постать. В простім, а про те дорогім і елегантнім домовім убраню, вона дуже живо занята тим, що „робить порядок“ у салонику: знимає полотняні футерали з мяхких, коштовних меблів і з золочених рам веркал та образів, уставлює сіметрично статуетки та оздобну посуду на комоді, придивлює ся і примірює, де би найкрасше стояти букетам з живих цвітів, що настромлені в делікатні вазоники з золоченого скла, розливають сильні пахощі на ввесь салоник. Упоравши ся з сим, підбігла до невеличкого, перламутром викладаного столика і накрутила старосвіцький металевий годинник, що довгий час без діла дрімав під хрусталевим клошем. Одним словом, молода пані „виганяє пустку“ з сего салоника, котрий, очевидно, чимало часу стояв пустий, запертий. В коминку тріщить і гуде веселий огонь, що звільна оживлює, огріває заморожене повітрє салоника, немов дострою є єго до оживлених рухів, цвитучого лиця і розіскрених очей пані дому.

Сторінки оригіналу видання 1897 року: титул і стор. 3.

присвяченого 150-річчю від дня народження Івана Франка (Львів, 27 вересня-1 жовтня 2006 р.).— Львів: Видавничий центр Львівського національного університету імені Івана Франка, 2010.— Т. 2.— С. 22.

[54] *Франко І.* Перехресні стежки: Повість.— Львів: Виданє ред. “Літературно-наукового вістника”, 1900.— 320 с. (на обкладинці вказано 1899 рік).

### 3.2. Принципи укладання словника

Принципи укладання частотних словників до творів І. Франка розроблено в межах проекту комплексного квантитативного опису творів І. Франка[55]. Згідно з ними вже укладено ЧС роману “Перехресні стежки”[56]. Роман “Для домашнього огнища” — другий за порядком твір письменника, лексика якого отримала комплексну кількісну характеристику у вигляді частотного словника.

ЧС роману “Для домашнього огнища” укладено на підставі лінгвостатистичного аналізу тексту твору. Він становить собою впорядкований список слів, забезпечений даними про частоту їх вживаності в тексті. Окремим словом вважаємо послідовність літер (тут апостроф і дефіс розглядаються як літера) між двома пропусками чи розділовими знаками, тому складні числівники виступали як різні слова. ЧС твору подає інформацію про словникові одиниці (леми) і про словоформи: парадигматичні форми і фонетичні варіанти слів.

Формування частотного словника здійснено за графічним збігом лем, і кожна частина мови мала свою схему об’єднання словоформ під лемою: аналогічну, як і в частотних словниках художньої прози[57], публіцистики[58], розмовно-побутового[59], наукового[60] та офіційно-ділового[61] стилів.

*Іменник* — до називного відмінка однини зведено форми всіх відмінків однини та множини. Частоту множинних іменників зведено до форми називного відмінка множини (*причандали, условини, зостанки*). Оскільки такі форми іменників як *домінікани, єзуїти, турки* і їм подібні позначають осіб і чоловічої, і жіночої статі, їх також не було зведено до чоловічого роду однини.

*Прикметник* — до називного відмінка однини чоловічого роду зведено відмінкові форми всіх родів в однині та множині, вищий і найвищий ступені порівняння, за винятком суплетивних форм, які зведено окремо до називного відмінка однини чоловічого роду вищого ступеня, наприклад, *більший, найбільший* зведено до *більший.* Така традиція встановилася у більшості словопокажчиків і частотних словників української мови.

*Займенник* — зведено відмінкові форми відповідно до типу відмінювання.

*Числівник* — зведено відмінкові форми відповідно до типу відмінювання.

*Дієслово* — зведено до інфінітива синтетичні форми часу (теперішній, минулий і майбутній), форми наказового способу і дієприслівник, а також неособові форми на -но, -то. Аналітичні форми часу ми вважали синтаксичними утвореннями, кожну складову яких зареєстровано як окреме слово.

*Дієприкметник* — до називного відмінка однини чоловічого роду зведено відмінкові форми всіх родів в однині та множині, оскільки розглядаємо його як різновид віддієслівного прикметника із властивими йому основними категоріями (рід, число, відмінок) та типовою синтаксичною роллю означення[62].

*Прислівник* — зведено вищий і найвищий ступені порівняння, за винятком суплетивних форм.

Лематизацію слів з частками *-но*, *-таки*, *-то*, слововживань з якими у тексті роману налічується кілька десятків, реалізовано так: **змінні** частини мови було зведено до початкових форм (*розповідж-но* до *розповісти*); у **незмінних** частинах мови ці частки збережено (*тим-то, якось-то, якби-то, все-таки, нарешті-таки, зовсім-таки…*).

---

55 *Бук С.* Квантитативна параметризація...

56 *Бук С., Ровенчак А.* Цит. праця.

57 Частотний словник сучасної української художньої прози / За ред. *Перебийніс В. С.*— К.: Наукова думка, 1981.— Т. 1.— 863 с.; Т. 2.— 855 с.

58 *Дарчук Н. П., Грязнухіна Т. О.* Цит. праця.

59 *Бук С.* 3 000 найчастотніших слів розмовно-побутового стилю сучасної української мови / Наук. ред. *Ф. С. Бацевич.*— Львів: ЛНУ імені Івана Франка, 2006.— 180 с.

60 *Бук С.* 3 000 найчастотніших слів наукового стилю сучасної української мови / Наук. ред. *Ф. С. Бацевич.*— Львів: ЛНУ імені Івана Франка, 2006.— 192 с.

61 *Бук С.* Частотний словник офіційно-ділового стилю: принципи укладання і частотні характеристики // Лінгвістичні студії.— 2006.— Вип. 14.— С 184–188.

62 *Вихованець І. Р.* Частиномовний статус дієприкметників // Українська мова.— 2003.— №2.— С. 54–60.

### 3.3. Фонетичні варіанти

До одної початкової форми зведено фонетичні варіанти слів, де чергування початкових чи кінцевих літер пов'язане з милозвучністю мови, а саме: дієслова з постфіксами *-ся / -сь*; сполучники *щоб / щоби, і / й; частки ж / же, б / би, ще/іще*; прийменники *у / в, з / із / зі / зо* та деякі інші; слова з відповідними префіксами (*вложити/уложити, весь / увесь / ввесь, всякий / усякий*), а також *вулиця / улиця, вухо / ухо.*

Не зводилися до єдиного слова такі: *влюблений/ улюблений, вклад/уклад, впадати/упадати*, оскільки вони мають різні значення:

... мов який **влюблений** хлопчина ... = *закоханий*
... не доторкнувся своєї **улюбленої** легоміни ... = *який подобається більше, ніж інші*

Твоя пенсія наразі буде йти на **вклади** в господарство = ***внески***
... цілим вихованням, цілим життям, **укладом**, ... попсовано її етичні основи = *встановлений порядок, що склався у житті, побуті, родині*

І чого ж тут гніватись і в пафос **впадати**? = ***пафосувати*** *(дія за значенням іменника)*
**Впадала** на різні здогади ... = ***здогадувалася*** *(дія за значенням іменника)*
Супроти рішучого зізнання сих дівчат усяке підозріння супроти вашої жінки **упадає**. = ***зникає***

В одній словниковій статті подано слова *ледве / ледво*, *трохи / троха, стільки / стілько* і подібні, оскільки у 50-томному виданні творів І. Франка авторську форму (*троха, стілько ...*) залишено у мовленні персонажів, а сучасну літературну форму (*трохи, стільки ...*) подано в інших випадках[63].

### 3.4. Омонімія

У словнику розмежовано лексичну та лексико-граматичну омонімію методом контекстуального аналізу: *а* (сполучник / частка / вигук), *раз* (прислівник / іменник / сполучник), *та* (займенник / сполучник / частка), *так* (частка / прислівник / сполучник), *то* (сполучник / частка / займенник), *волів* (іменник / дієслово), *дні* (Н.в. мн. від *день* і Р.в. одн. від *дно*) тощо. В омографах подано наголос: *мукá* і *мýка, вúкликати* і *викликáти* тощо.

У ЧС для розрізнення омонімів у дужках подано або вказівку на значення (*нічóго* (присудк.сл.)), або на частиномовну належність *(мати* (*ім.*) і *мати* (*дієсл.*)).

У додатку подано першу сотню найчастотніших лем (слів, зведених до словникової форми). Очікувано високий ранг мають, окрім службових частин мови, назви персонажів *капітан, Анеля, Редліх, Юлія*, дієслова *бути, могти, мати, знати*, іменників *пан, пані, рука, діти*.

### 4. Деякі результати статистичних підрахунків

1. Обсяг роману (*N*) — 44 840 слововживань.
2. Кількість різних словоформ ($V_ф$) — 11 505,
3. Кількість різних слів (*V*) — 6486.
4. Багатство словника (індекс різноманітності), тобто відношення обсягу словника лексем до обсягу тексту (*V/N*), становить 0,145.
5. Середня повторюваність слова в тексті, тобто відношення обсягу тексту до обсягу словника лексем (*N/V*) — 6,9. Іншими словами, кожне слово в середньому вжито в досліджуваному тексті приблизно 7 разів.
6. Кількість hapax legomena — слів із частотою 1 — ($V_1$ =3363) складає 7,5% тексту і 51,85% (більше за половину!) словника.

---

63 Від редакційної колегії // *Франко І.* Зібрання творів у 50-ти томах.— Т. 1.— К.: Наукова думка, 1979.— С. 14–15.

7. Індекс винятковості у тексті, тобто відношення $V_1$ до обсягу тексту ($V_1/N$) 0,075.

8. Індекс винятковості у словнику, тобто відношення $V_1$ до обсягу словника ($V_1/V$), становить 0,52. Два останні числа — показники варіативності лексики.

9. Протилежним до індексу винятковості є індекс концентрації — відношення кількості слів з високою частотою (10 і вище, їх у тексті $N_{10}$ = 32 516, у словнику — $V_{10}$ = 598) до обсягу тексту ($N_{10}/N$ =0,752) або словника ($V_{10}/V$ =0,092).

Відносно невелика кількість високочастотної лексики (і, відповідно, низький індекс концентрації) та порівняно велика кількість слів із частотою 1 (і, відповідно, високий індекс винятковості) свідчать про неабияке різноманіття лексики роману.

Цікаво порівняти кількісні дані двох романів І. Франка: “Для домашнього огнища” та “Перехресні стежки” на основі їх ЧС. Зіставлення характеристик різних текстів має сенс лише у випадку приблизно однакової кількості слововживань в аналізованих матеріалах, оскільки збільшення обсягу словника відбувається не пропорційно до збільшення обсягу тексту, а дещо повільніше[64]. Оскільки обсяг роману “Для домашнього огнища” більш як удвічі менший за обсяг “Перехресних стежок”, то, наприклад, вищий індекс різноманітності його лексики чи більша частка hapax legomena у словнику є очікуваними явищами. Ця проблема потребує окремого детальнішого дослідження. У табл. 1 наведено зіставлення деяких параметрів двох творів. Для коректного порівняння подано результати розрахунків на матеріалі перших 44 840 слововживань роману “Перехресні стежки”, що відповідає обсягові роману “Для домашнього огнища”.

Таблиця 1.

**Зіставлення кількісних характеристик романів І. Франка**
**“Для домашнього огнища” та “Перехресні стежки”**

| **Кількісні характеристики** | **Для домашнього огнища** | **Перехресні стежки (*перші 44 840 слововживань*)** | **Перехресні стежки (*весь текст*)** |
|---|---|---|---|
| Обсяг роману ($N$) | 44 840 | 44 840 | 93 888 |
| Кількість різних словоформ ($V_ф$) | 11 505 | 11 506 | 19 448 |
| Кількість різних слів ($V$) | 6486 | 6561 | 9978 |
| Багатство словника (індекс різноманітності) | 0,145 | 0,146 | 0.106 |
| Середня повторюваність слова в тексті ($N/V$) | 6,9 | 6,8 | 9.41 |
| Кількість hapax legomena ($V_1$) | 3363 | 3586 | 4907 |
| Індекс винятковості у тексті ($V_1/N$) | 0,075 | 0,080 | 0.0523 |
| Індекс винятковості у словнику ($V_1/V$) | 0,519 | 0,547 | 0.492 |
| Індекс концентрації тексту ($V_{10}/N$) | 0,752 | 0,731 | 0.795 |
| Індекс концентрації словника ($V_{10}/V$) | 0,092 | 0,087 | 0.113 |

Як видно із таблиці, для текстів І. Франка однакового розміру основні квантитативні показники фактично збігаються, а саме: кількість різних словоформ та слів, багатство словника, середня повторюваність слова в тексті тощо. Дещо відрізняються лише кількість hapax legomena (і через це — індекс винятковості у тексті та словнику), а також індекс концентрації тексту і словника.

[64] Частотний словник сучасної української художньої прози / За ред. *Перебийніс В. С.*— К.: Наукова думка, 1981.— Т. 1.— С. 14; *Тулдава Ю. П.* Проблемы и методы ... — С. 62-64.

Визначальною ознакою ЧС є те, що у ньому слова розміщують у порядку спадання частоти, тобто слово з найбільшою частотою має ранг 1, наступне — 2 і т. д. Таке подання інформації дає змогу обчислити, яку частку тексту (покриття) становлять слова з найбільшою частотою. Величину покриття тексту для певного ранґу обчислюють як відношення суми абсолютних частот усіх слів з меншими ранґами до загальної кількості слів у тексті. Співвідношення між ранґом слова та покриттям тексту подано у таблиці (*Табл.* 2) і на рисунку (*Рис.* 1).

Таблиця 2.

**Співвідношення між ранґом слова та величиною покриття тексту**

| Ранґ | Покриття | Ранґ | Покриття | Ранґ | Покриття |
|---|---|---|---|---|---|
| 1 | 3,33 | 200 | 58,83 | 1500 | 83,99 |
| 5 | 11,78 | 300 | 63,87 | 2000 | 87,39 |
| 10 | 18,88 | 400 | 67,45 | 3000 | 91,96 |
| 25 | 31,13 | 500 | 70,25 | 4000 | 94,46 |
| 50 | 40,19 | 600 | 72,56 | 5000 | 96,69 |
| 75 | 45,65 | 750 | 75,39 | 6000 | 98,92 |
| 100 | 49,59 | 1000 | 79,01 | 6484 | 100,00 |

Із Таблиці 2 видно, що перших за частотністю 25 слів покривають 31% тексту, перших 100 слів — 50%, 1000 слів — 79% тексту.

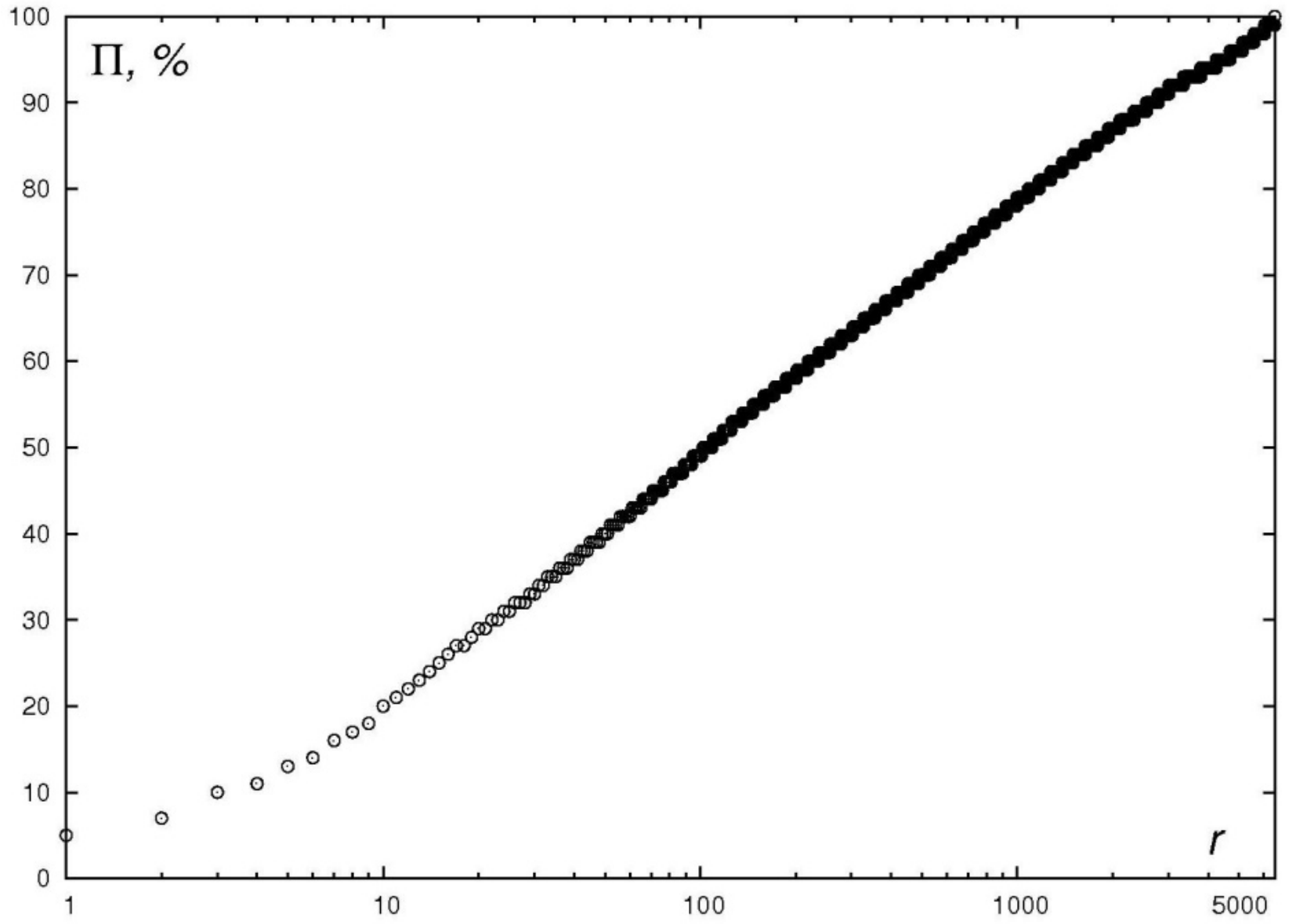

Рис. 1. Залежність величини покриття тексту (П) від ранґу леми (*r*).

**5. Висновки**

Роман "Для домашнього огнища" — важливий твір у творчості Івана Франка. Квантитативний опис його лексики у цій статті здійснено вперше. ЧС аналізованого роману укладено за методикою, що розроблена у проекті комплексного квантитативного опису творів І. Франка. ЧС роману І. Франка "Для домашнього огнища", за класифікацією частотних словників, наведеною В. Перебийніс[65], є 1) за одиницями підрахунку — частотний словник слів та словоформ; 2) за обсягом вибірки — невеликий; 3) за характером вибірки — частотний словник конкретного твору; 4) за обсягом — повний; 5) за характером подання матеріалу — алфавітно-частотний і частотно-алфавітний; 6) за статистичними характеристиками одиниць — частотний словник абсолютної та відносної частоти без статистичних її оцінок.

[65] *Перебийніс В. С.* Частотний словник // Українська мова: Енциклопедія / редкол.: *В. М. Русанівський* та інші.— К.: Українська енциклопедія, 2000.— С. 790.

ЧС роману дав змогу отримати важливі статистичні характеристики тексту, такі як обсяг роману, кількість різних словоформ та слів, індекс різноманітності (багатство словника), середня повторюваність слова в тексті, кількість hapax legomena, індекс винятковості у тексті та у словнику, індекс концентрації, співвідношення між рангом слова та величиною покриття тексту.

Крізь призму ЧС лексичний склад роману отримує нову інтерпретацію.

(Львів)

## IVAN FRANKO'S NOVEL *Dlja domashnjoho ohnyshcha (For the Hearth)* IN THE LIGHT OF THE FREQUENCY DICTIONARY

In the article, the methodology and the principles of the compilation of the Frequency dictionary for Ivan Franko's novel *Dlja domashnjoho ohnyshcha (For the Hearth)* are described. The following statistical parameters of the novel vocabulary are obtained: variety, exclusiveness, concentration indexes, correlation between word rank and text coverage, etc. The main quantitative characteristics of Franko's novels *Perekhresni stezhky (The Cross-Paths)* and *Dlja domashnjoho ohnyshcha* are compared on the basis of their frequency dictionaries.

Key words: word frequency; frequency dictionary; rank; number of the word occurrences; hapax legomena; variety, exclusiveness, concentration indexes; text coverage.

### **Додаток.**
### **Перша сотня найчастотніших лем роману І. Франка “Для домашнього огнища”**

| Ранг | Лема | Абс. част. | Відн. част., % | Покр., % |
|---|---|---|---|---|
| 1 | І—1393; Й—101 | 1494 | 3,33 | 3,33 |
| | *-СЯ* | *1285* | *2,87* | |
| 2 | ВІН | 1020 | 2,27 | 8,47 |
| 3 | НЕ | 1000 | 2,23 | 10,70 |
| 4 | В—728;У—241 | 969 | 2,16 | 12,86 |
| 5 | Я | 800 | 1,78 | 14,65 |
| 6 | З—620; ІЗ—54; ЗІ—33; ЗО—13 | 720 | 1,61 | 16,25 |
| 7 | НА | 703 | 1,57 | 17,82 |
| 8 | ТОЙ | 589 | 1,31 | 19,13 |
| 9 | СЕЙ | 588 | 1,31 | 20,45 |
| 10 | БУТИ | 585 | 1,30 | 21,75 |
| 11 | ЩО(займ.) | 577 | 1,29 | 23,04 |
| 12 | ДО | 543 | 1,21 | 24,25 |
| 13 | ВОНА | 529 | 1,18 | 25,43 |
| 14 | ЩО(спол.) | 529 | 1,18 | 26,61 |
| 15 | КАПІТАН | 489 | 1,09 | 27,70 |
| 16 | А(спол.) | 478 | 1,07 | 28,76 |
| 17 | ТИ | 361 | 0,81 | 29,57 |
| 18 | СВІЙ | 276 | 0,62 | 30,19 |
| 19 | Ж—245; ЖЕ—25 | 270 | 0,60 | 30,79 |
| 20 | ВОНИ | 255 | 0,57 | 31,36 |
| 21 | МОГТИ | 246 | 0,55 | 31,90 |
| 22 | ВЕСЬ—154; УВЕСЬ—87; ВВЕСЬ—1 | 242 | 0,54 | 32,44 |
| 23 | АНЕЛЯ(ім'я) | 236 | 0,53 | 32,97 |
| 24 | ТАКИЙ | 233 | 0,52 | 33,49 |

| Ранг | Лема | Абс. част. | Відн. част., % | Покр., % |
|---|---|---|---|---|
| 25 | ТА(спол.) | 226 | 0,50 | 33,99 |
| 26 | ЗНАТИ | 206 | 0,46 | 34,45 |
| 27 | ЗА | 202 | 0,45 | 34,90 |
| 28 | КОЛИ | 202 | 0,45 | 35,35 |
| 29 | БИ—162; Б—39 | 201 | 0,45 | 35,80 |
| 30 | ПРО | 198 | 0,44 | 36,24 |
| 31 | МАТИ(дієсл.) | 188 | 0,42 | 36,66 |
| 32 | ЯКИЙСЬ | 186 | 0,41 | 37,08 |
| 33 | СЕБЕ | 183 | 0,41 | 37,49 |
| 34 | МІЙ | 171 | 0,38 | 37,87 |
| 35 | ВЖЕ—120; УЖЕ—49 | 169 | 0,38 | 38,24 |
| 36 | АЛЕ | 168 | 0,37 | 38,62 |
| 37 | ТАК(присл.) | 164 | 0,37 | 38,99 |
| 38 | МОВИТИ | 161 | 0,36 | 39,34 |
| 39 | ЩЕ—143; ІЩЕ—18 | 161 | 0,36 | 39,70 |
| 40 | ЯКИЙ | 154 | 0,34 | 40,05 |
| 41 | МИ | 148 | 0,33 | 40,38 |
| 42 | ПО | 144 | 0,32 | 40,70 |
| 43 | САМ | 137 | 0,31 | 41,00 |
| 44 | ЯК(спол.) | 136 | 0,30 | 41,31 |
| 45 | ПАН | 135 | 0,30 | 41,61 |
| 46 | ВІД | 134 | 0,30 | 41,91 |
| 47 | ЩОБИ—110; ЩОБ—24 | 134 | 0,30 | 42,21 |
| 48 | ТУТ | 130 | 0,29 | 42,50 |
| 49 | ОДИН | 126 | 0,28 | 42,78 |

| Ранґ | Лема | Абс. част. | Відн. част., % | Покр., % |
|---|---|---|---|---|
| 50 | РУКА | 124 | 0,28 | 43,05 |
| 51 | ТЕПЕР | 124 | 0,28 | 43,33 |
| 52 | ТАМ | 122 | 0,27 | 43,60 |
| 53 | ХВИЛЯ | 121 | 0,27 | 43,87 |
| 54 | ОКО | 118 | 0,26 | 44,13 |
| 55 | РАЗ(ім.) | 112 | 0,25 | 44,38 |
| 56 | НІ | 111 | 0,25 | 44,63 |
| 57 | ДІТИ(ім.) | 103 | 0,23 | 44,86 |
| 58 | РЕДЛІХ(прізв.) | 103 | 0,23 | 45,09 |
| 59 | ІТИ—42; ЙТИ—59 | 101 | 0,23 | 45,32 |
| 60 | ВИ | 100 | 0,22 | 45,54 |
| 61 | ПЕРЕД(прийм.) | 100 | 0,22 | 45,76 |
| 62 | ХОТІТИ | 100 | 0,22 | 45,99 |
| 63 | ЖІНКА | 99 | 0,22 | 46,21 |
| 64 | ДЛЯ | 97 | 0,22 | 46,42 |
| 65 | СЛОВО | 94 | 0,21 | 46,63 |
| 66 | АДЖЕ(част.) | 93 | 0,21 | 46,84 |
| 67 | КОТРИЙ | 91 | 0,20 | 47,04 |
| 68 | ТО(спол.) | 88 | 0,20 | 47,24 |
| 69 | ЛИЦЕ | 86 | 0,19 | 47,43 |
| 70 | СКАЗАТИ | 86 | 0,19 | 47,62 |
| 71 | НУ | 84 | 0,19 | 47,81 |
| 72 | ДУША | 81 | 0,18 | 47,99 |
| 73 | ГОЛОВА | 79 | 0,18 | 48,17 |
| 74 | БАЧИТИ | 78 | 0,17 | 48,34 |
| 75 | МУСИТИ | 78 | 0,17 | 48,51 |
| 76 | НІЩО | 78 | 0,17 | 48,69 |

| Ранґ | Лема | Абс. част. | Відн. част., % | Покр., % |
|---|---|---|---|---|
| 77 | ТІЛЬКИ | 77 | 0,17 | 48,86 |
| 78 | ЩОСЬ(займ.) | 77 | 0,17 | 49,03 |
| 79 | ТО(част.) | 76 | 0,17 | 49,20 |
| 80 | ЧИ(част.) | 75 | 0,17 | 49,37 |
| 81 | НАВІТЬ | 74 | 0,17 | 49,53 |
| 82 | ПАНІ | 74 | 0,17 | 49,70 |
| 83 | ЗОВСІМ | 73 | 0,16 | 49,86 |
| 84 | ПРИ | 73 | 0,16 | 50,02 |
| 85 | АНІ | 72 | 0,16 | 50,19 |
| 86 | БЕЗ | 71 | 0,16 | 50,34 |
| 87 | ДУЖЕ | 71 | 0,16 | 50,50 |
| 88 | ЮЛІЯ(ім'я) | 71 | 0,16 | 50,66 |
| 89 | КІЛЬКА | 70 | 0,16 | 50,82 |
| 90 | ГОВОРИТИ | 69 | 0,15 | 50,97 |
| 91 | МОЖЕ | 69 | 0,15 | 51,12 |
| 92 | СКРИКНУТИ | 69 | 0,15 | 51,28 |
| 93 | АБО(спол.) | 68 | 0,15 | 51,43 |
| 94 | МОВ | 68 | 0,15 | 51,58 |
| 95 | ДУМКА | 67 | 0,15 | 51,73 |
| 96 | ТІЛЬКО | 66 | 0,15 | 51,88 |
| 97 | ГОЛОС | 65 | 0,14 | 52,02 |
| 98 | ТАК(част.) | 65 | 0,14 | 52,17 |
| 99 | ХТО | 65 | 0,14 | 52,31 |
| 100 | ЧАС | 65 | 0,14 | 52,46 |
| 101 | ЗНОВ | 64 | 0,14 | 52,60 |
| 102 | ТВІЙ | 64 | 0,14 | 52,74 |
| 103 | ЧИ(спол.) | 64 | 0,14 | 52,89 |